\documentclass[conference]{IEEEtran}
\IEEEoverridecommandlockouts
\usepackage{multirow} 
\usepackage{url}
\usepackage{amsmath,amssymb,amsfonts}
\usepackage[numbers,sort&compress]{natbib}
\usepackage{graphicx}
\usepackage{textcomp}
\usepackage{xcolor}
\usepackage{subcaption}
\usepackage{stfloats}
\usepackage{bm}
\usepackage{balance}
\usepackage[caption=false,font=footnotesize]{subfig}
\usepackage[margin=1in]{geometry}
\usepackage{amsmath,amssymb,amsthm}
\usepackage{mathtools}
\usepackage{enumitem}
\usepackage{xcolor}
\usepackage{parskip}
\usepackage{algorithm}
\usepackage{algpseudocode}
\usepackage{float}
\usepackage{booktabs}
\usepackage{tabularx}
\usepackage[dvipsnames]{xcolor}

\newcommand{\pvec}{\vec{p}}
\newcommand{\vvec}{\vec{v}}
\newcommand{\fvec}{\mathbf{f}}
\newcommand{\qvec}{\mathbf{q}}
\newcommand{\wvec}{\boldsymbol{\omega}}
\newcommand{\Jmat}{\mathbf{J}}
\definecolor{mauve}{HTML}{9A4EAE}
\definecolor{burntorange}{HTML}{CC5500}

\newcommand{\skewm}[1]{{[#1]^{\times}}}

\def\BibTeX{{\rm B\kern-.05em{\sc i\kern-.025em b}\kern-.08em
    T\kern-.1667em\lower.7ex\hbox{E}\kern-.125emX}}

\begin{document}

\title{Illumination-Aware Active Perception for\\ Spacecraft Inspection}

\author{
\IEEEauthorblockN{Sagarika Rao Valluri}
\IEEEauthorblockA{\textit{Mechanical and Aerospace Engineering} \\
\textit{New York University}}
\and
\IEEEauthorblockN{Benjamin Riviere}
\IEEEauthorblockA{\textit{Mechanical and Aerospace Engineering} \\
\textit{and Computer Science} \\
\textit{New York University}}
}
\maketitle

\begin{abstract}
Harsh and rapidly varying orbital illumination introduces failure modes during spacecraft inspection. Existing planners rely on line-of-sight inspection metrics which perform occlusion checks against a known target and conflate geometric visibility with task-relevant information. 
These metrics overlook illumination effects including eclipse, self-shadowing, and specular reflection from spacecraft materials and tend to declare components as inspected despite the corresponding images lacking sufficient information.
We propose an inspection-planning framework that replaces geometric visibility with a learned measure of camera information derived from a classifier variational autoencoder (VAE) that operates directly on the captured image.
The encoder model is trained to classify spacecraft surface components from rendered patches, and its total latent variance is used to learn the varying illumination of said components, replacing binary visibility checks with a continuous, image-grounded quantity. 
We evaluate the approach by implementing a Guidance, Navigation, and Control architecture in a photorealistic physics simulator, and vary lighting and viewing geometries with multiple orbits. We show that, although all approaches successfully maximize their own objective, our approach results in better illumination: a mean brightness of $0.57$, with $81.6\%$ of observations above $0.25$, compared with $0.02$ ($3.0\%$) for geometric planning with an illumination band and $0.04$ ($6.0\%$) for purely geometric planning.

\end{abstract}

\begin{IEEEkeywords}
spacecraft inspection, active perception, variational autoencoder, proximity operations, information-driven guidance
\end{IEEEkeywords}

\section{Introduction}
Autonomous inspection of resident space objects is essential for space situational awareness and is an important tool to execute proximity operations in constrained orbital environments. Inspection is necessary in missions like debris removal, satellite repair~\cite{AILOR2022449}, and in-orbit assembly and manufacturing. Recent missions include the Remora mission conducted by Starfish Space and Impulse Space ~\cite{starfish2025}, Qingzhou prototype spacecraft and Astroscale's ADRAS‑J, a demonstration satellite that successfully captured close range pictures of space debris~\cite{astroscale2026}.

\begin{figure}[t]
    \centering
    \includegraphics[width=\columnwidth]{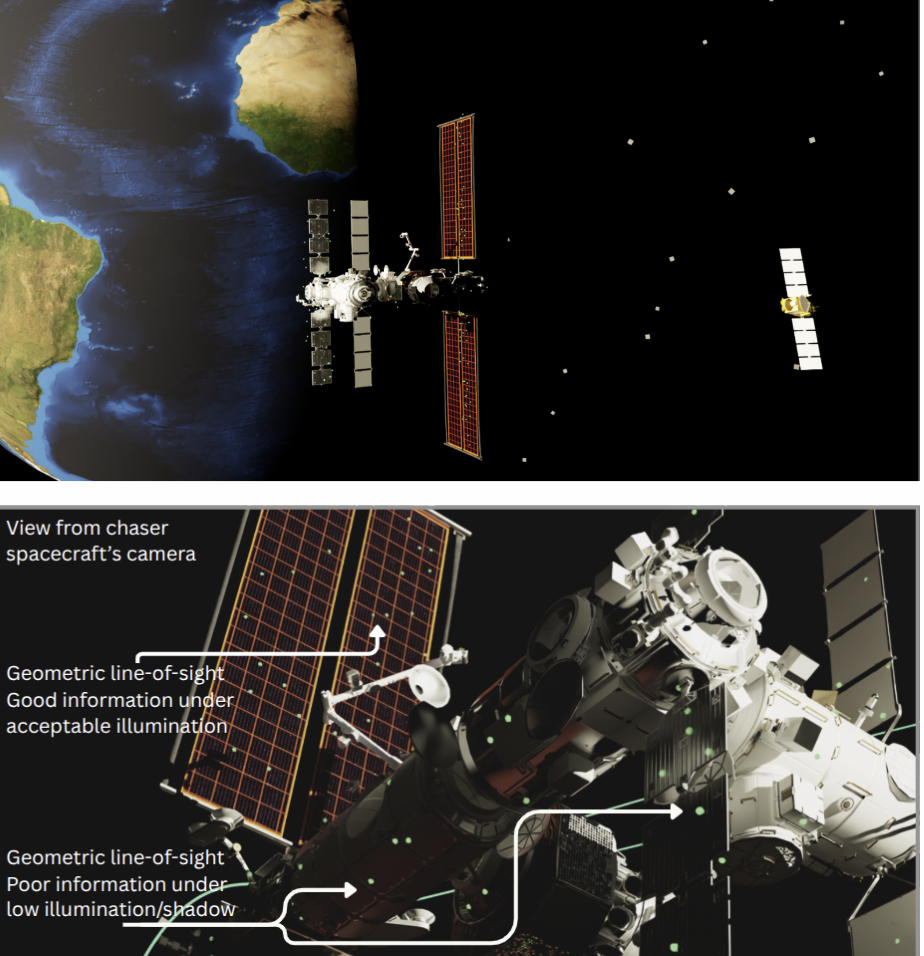}
    \caption{Active-perception inspection conducted by the chaser to gain maximum information under illumination constraints. Top: chaser spacecraft on the right, target spacecraft on the left. Bottom: chaser's camera view of target spacecraft. Green dots are Points of Interest (POI) successfully inspected.}
    \label{fig:conops}
\end{figure}
The primary objective of autonomous inspection is to maximize information gain of the target spacecraft in motion. This task comprises of gaining information about the target pose (in orbit, tumbling, undefined motion etc) and of specific component state on the target. Extensive research has been conducted by estimating target's (relative) pose for navigation of non-cooperative targets~\cite{fourie2014} for single agent and multi agent inspection tasks~\cite{sabatini2020}. These models have used camera data to train vision models to estimate pose by using navigation estimates to ensure target body is within the field of view. Given a target's pose (inferred through above methods) autonomous methods are used to inspect state of components on the target using measurements like encounter geometry-incidence angle, range, visibility mask, and analytical models of illumination often amounting to a visibility check. Often, hierarchical models~\cite{nakka2022} are built with a Guidance, Navigation, and Control (GNC) stack that explicitly sets an information metric to aid coverage and mapping of the target spacecraft. 
Ultimately, the objective is to identify components by determining whether a surface element is a solar array, a radiator, or a docking interface, and if its condition is nominal or degraded. Without camera data, geometrical checks limit the planner's ability to distinguish orbital viewpoints that yield photometrically informative measurements from those that do not, regularly admitting false positives/negatives during the inspection task.

To address this problem, we replace the geometrically computed metric with a learned metric that operates on image pixels directly from the onboard camera. Specifically, we train a classifier variational auto-encoder (VAE)~\cite{kingma} to classify target components from rendered imagery and learn corresponding illumination distributions. 
During evaluation, the model's latent variance is used for our inspection probability where it is high for dark, occluded, specular, or out-of-distribution patches and is low for well-resolved, in-distribution patches. To validate our solution, we implement an inspection-based Guidance, Navigation, and Control stack in the IsaacSim~\cite{NVIDIA_Isaac_Sim} a photorealistic simulator, and we compare closed loop performance of our learned metric to geometric and analytic baselines. 
We find that the closed-loop system equipped with our learned model captures images with a mean brightness of $0.57$, compared with $0.04$ and $0.02$ for the geometric and illumination-band baselines, with $81.6\%$ of observations above $0.25$ versus $6.0\%$ and $3.0\%$ (an increase of approximately $76$ percentage points over the stronger baseline). We also find that its policy nearly matches the illumination-band baseline on that baseline's own reward ($193$ versus $197$ POI's), while the reverse does not hold ($100$ versus $190$ under our reward).
\section{Related Work}
\label{related_work}
Inspection planners comprise of two coupled components: a guidance system that determines the viewpoints to visit, and a metric that quantifies whether points of interest (POI) from said viewpoints are considered inspected.

The metric informs viewpoint selection, and guidance executes it, and guidance aids the goal of achieving complete coverage (inspecting all POI's under constraints). 
Within aerospace literature, guidance methods can be broadly classified into three lines of work:

\textit{Trajectory-based inspection:} These methods optimize trajectories to maximize an inspection metric under fuel and safety constraints, often formulated with analytical checks.
The first approach~\cite{nakka2022} uses passive relative orbits (PROs) as the primitive of trajectory design and plans fuel-optimal reconfiguration. The inspection metric is a cost function that is minimized when POI's are considered inspected (a line of sight check based on sensor distance).  
The second approach~\cite{malyuta} directly optimizes trajectories as an optimization over finite thrust maneuvers subject to constraints solved as a convex program. 
The third approach~\cite{lauinger} decomposes the target into primitive shapes and projects Lissajous, circling, or helical patterns onto the surface, extending to constellations as a multi-rendezvous sequencing problem.

\textit{Learning-based inspection:} Reinforcement learning (RL) is an alternative optimization that trains a policy function to maximize coverage and minimize fuel. 
A multi agent deep RL approach~\cite{aurand2023} inspects a tumbling, non convex target (a point cloud satellite model) using three decentralized agents that select view points to maximize information gain under a fuel penalty. A point is considered inspected if it falls under the sensor's field of view cone and is unoccluded. This was extended to a scalable multi-agent setting~\cite{dunlap20242024} and a six-degree-of-freedom task with attitude control under runtime assurance safety constraints~\cite{dunlap2024uav}. Here, the target is represented by 100 inspection points equally distributed across the surface of a 10m radius sphere and the condition for inspection is also geometric where the point must be in a field of view and within the illumination gate specified. Although illumination is considered, the model abstracts the target spacecraft to a uniform shape such as, a sphere. 

\textit{Safety constrained inspection:} Deploying learned policies on real spacecraft is contingent on formal safety guarantees, commonly provided by control barrier functions (CBFs) that enforce constraints. Recent work~\cite{dunlap2024uav} uses a discrete-CBF safety filter to enforce position, velocity, attitude, and thermal constraints on an RL inspection agent. The paper uses the  Blinn-Phong (analytical) illumination model by evaluating each point's illumination through a shadow gate (requiring a positive inner product between the surface normal and the Sun direction) and excluding out-of-band points from the reward. Subsequent work extends CBF safety to six-degree-of-freedom inspection~\cite{ravaioli2023} to multi-spacecraft close proximity inspection with collision, occlusion, and sunlight avoidance~\cite{shi2025}. A meta-reinforcement learning robust model incorporates CBF's~\cite{wijayatunga2026} and considers the sun's angle as part of the policy. 

Across all planners the inspection metric relies on functions which are analytically computed and illumination, when considered is geometrical. This limits models to only a line-of-sight check and assumes that is an adequate measure for inspection. In Sun synchronous orbits~\cite{stephenson2025}, surfaces experience disproportionate illumination conditions and parts of the target are always in shadow or dimly lit and even with an analytical band these classical inspection models will be limited.
\subsection{Inspection models} 
Within robotics, different vision models have been employed for inspection and mapping tasks. Volumetric information-gain methods \citep{isler2016information,delmerico2018comparison} use a probabilistic occupancy map and select the next-best view (NBV) to maximize expected entropy reduction (information metric). Learned NBV approaches replace the analytical gain with a network that predicts the best next pose from a partial point cloud \citep{mendoza2020nbvnet,zeng2020pcnbv} or a reinforcement-learning policy \citep{peralta2020nbvpolicy}. More recently scene representations have made it possible to estimate view utility directly in image space: ActiveNeRF \citep{pan2022activenerf}, NeU-NBV \citep{jin2023neunbv}, NeurAR \citep{ran2023neurar}, and uncertainty guided policy planners \cite{lee2022uncertainty} rank candidate views by predicted rendering uncertainty, while FisherRF \citep{jiang2023fisherrf} uses the Fisher information of the radiance field parameters as a view selection criterion. Perception-aware planners \citep{costante2018perception,zhang2019fisher} choose trajectories that keep trackable features in view, improving the robot's localization rather than its observation of a target.
These approaches use camera data, but for reconstruction and rendering rather than the measurement requirements to inspect targets. Under challenging orbital illumination, inspection metrics must instead be derived from the camera data itself similar to the models above. Currently, no existing planner combines a known target model with camera-informed inspection metrics under orbital illumination and this gap is addressed. 

\section{Problem Statement}

The inspection task is formulated as a Markov Decision Process (MDP)~\cite{puterman} as $M = \{S,A,T,R\}$ where $S$ corresponds to the state space, $A$ to the action space, $T$ is the transition kernel and $R$ is the reward function. No discount factor is used in determining the reward. We use the subscript $j$ for the point of interest (POI) index.

\textit{State Space}: The complete state $s$ is a combination of the physical state $x$ and inspection state $z$: $s =[x;z]$. The physical state $x \in \mathbb{R}^{n}$ of the system is the generalized position and velocity of the target and chaser, given as $x=[q_c;\dot{q}_c;q_t;\dot{q}_t]$ where every body $i \in \{c,t\}$ has position $q_i = [p_i, R_i]$ and velocity $\dot{q}_i =[v_i, \omega_i]$ where $p_i \in \mathbb{R}^{3}$, $R_i \in SO(3)$, $v_i \in \mathbb{R}^{3}$ and $\omega_i \in \mathbb{R}^{3}$, making the physical state dimension $n=24$.
The inspection state $z \in \{0,1\}^m$ is the imaging indicator variable determining if a POI is successfully observed, where $m$ is the total number of POIs assigned for the inspection task. 

\textit{Action Space}: 
The action $a$ consists of the force and torque of the chaser spacecraft: $a=[\fvec;\boldsymbol{\tau}]$, where $\fvec \in \mathbb{R}^3$ is the external force applied to the chaser by its thrusters and $\boldsymbol{\tau} \in \mathbb{R}^3$ is the external torque applied by the reaction wheels through a control allocation matrix, expressed in the chaser body frame.

\textit{Transition}: We assume a factorized model of the transition, in that the next physical state is independent of inspection state and the inspection state is updated from the current physical state alone: 
\begin{equation}
    p(s'|s,a) = p(x',z'|x,z,a) = p(x'|x,a)\,p(z'|z,x)
\end{equation}

The physical state transition $p(x'|x,a)$ is a time-discretization of standard mechanical equations for a target in an unactuated orbit. In particular, the relative translational dynamics are the two body relative motion linearized about the target's (circular) orbit~\cite{clohessy1960}, expressed in the target centered local vertical local horizontal frame, while the attitude of each body evolves on $SO(3)$ under Euler's rigid body equations. 

\begin{align}
    M(q_c) \ddot{q}_c + C(q_c,\dot{q}_c) + G(q_c) &= a \\ 
    M(q_t) \ddot{q}_t + C(q_t,\dot{q}_t) + G(q_t) &= 0 
\end{align}                             
where $M$ is the mass matrix, $C$ contains the Coriolis terms, $G$ contains the gravity terms, and the chaser is actuated by action $a$. 

The inspection state's transition $p(z'|z,x)$ is:

    \begin{align}
        p(z'_j|z_j,x) =
        \begin{cases}
            1 & z'_j=1,\ z_j=1\\
            \nu_j(x) & z'_j=1,\ z_j=0\\
            1-\nu_j(x) & z'_j=0,\ z_j=0
        \end{cases}
    \end{align}

where $\nu_j(x)$ is the probability of detecting POI $j$ from physical state $x$. We instantiate different inspection algorithms based on the choice of $\nu$, specified in Sec.~\ref{sec:method}.

\textit{Reward}: The reward $R$ is the expected change in detection of all the POIs: 
\begin{align}
\label{reward}
    R(s,a) = \mathbb{E}_{s' \sim p(\cdot \mid s,a)}\left[\sum_{j=1}^{m} (z_j' - z_j)\right].
\end{align}
\section{Guidance, Navigation, and Control (GN\&C) System Architecture}
\label{sec:gnc}
\begin{figure}[t]
\centering
\includegraphics[width=\columnwidth]{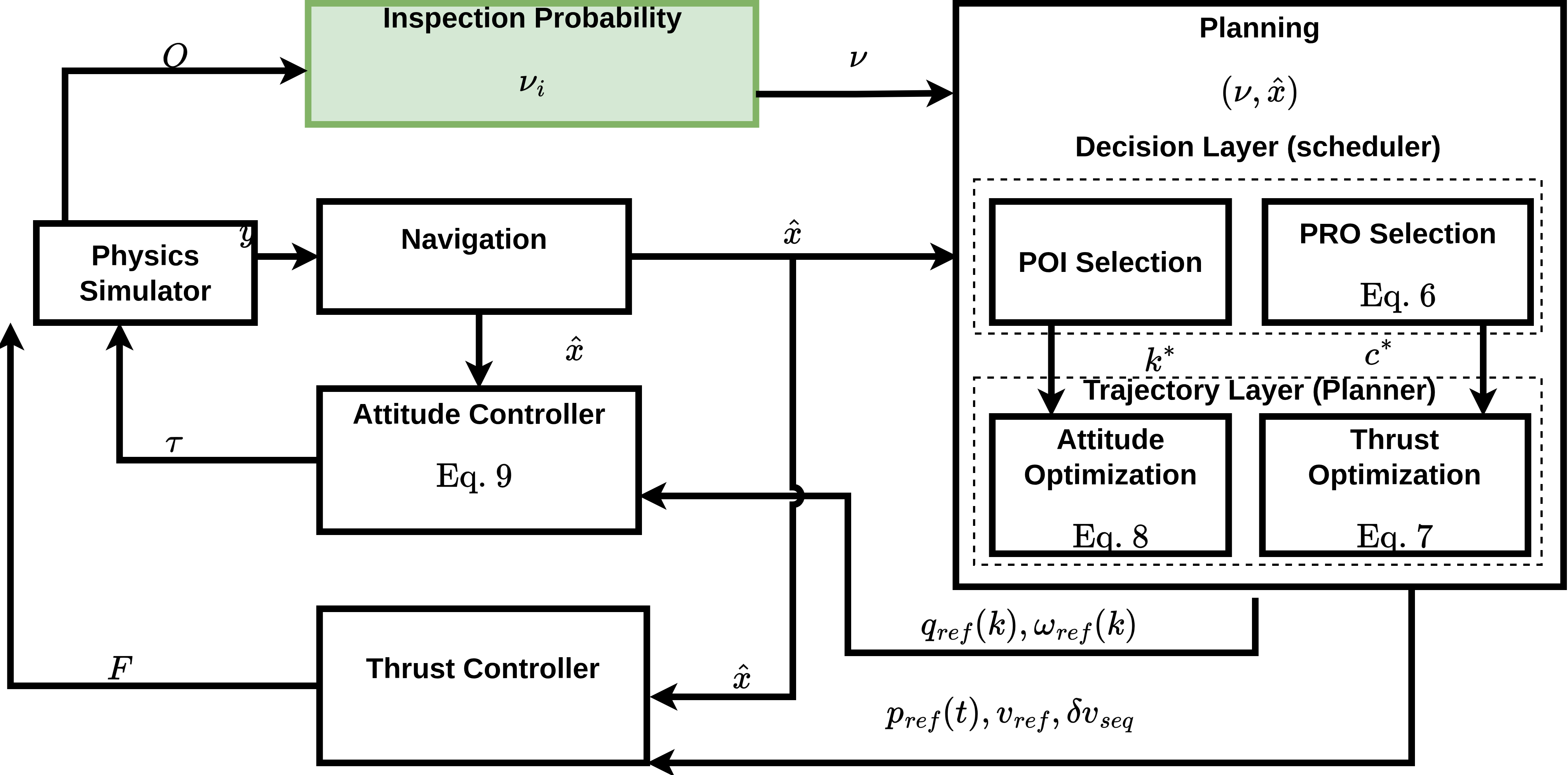}
\caption {Closed-loop inspection architecture. Only the inspection metric (shown in green) differs between the three models; the decision layer, trajectory layer, and controllers are common to all three.}
\label{fig:arch}
\end{figure}

The GN\&C system architecture is shown in (Fig.~\ref{fig:arch}): navigation, discrete planning, continuous planning, attitude control, thrust control, and inspection metric and are discussed in the section below as part of a closed loop model.

\textit{Navigation} The navigation model produces the perfect physical state $x = [q_c; \dot{q}_c; q_t; \dot{q}_t]$ and all quantities are directly read from the simulator with zero measurement noise for chaser and target. Given perfect physical state we realize a hierarchical planning algorithm.

\textit{Discrete Planning.}
The decision layer operates on the inspection state $z$ and the per-POI
detection probabilities $\nu_j$, and produces two discrete targets at two rates: at the low rate, a passive relative orbit (PRO) for the chaser to occupy; at the mid rate, a POI for the camera to point at. Both selections maximize the same quantity, the
expected reward of Eq.~\ref{reward}. Every candidate in the PRO pool $\mathcal{C}$ is rolled out over one period with the prediction model, giving $K$ predicted chaser positions per candidate. At each position $k$ the camera is pointed greedily at POI $e^{c}_k$, and the inspection probability $\nu$ gives the detection probability $\nu^{c}_{j,k}$ of each POI $j$. The candidate that maximizes the expected number of newly inspected POIs is selected,
\begin{equation}
c^\star=\arg\max_{c\in\mathcal{C}}\;
\sum_{j=1}^{m}(1-z_j)\Bigl[\,1-\prod_{k=1}^{K}\bigl(1-\nu^{c}_{j,k}\bigr)\Bigr],
\label{pro_select}
\end{equation}
where $z_j\in\{0,1\}$ indicates whether POI $j$ has already been inspected and $m$ is the number of POIs. The bracketed term is the probability that POI $j$ is detected at least once along the roll-out, so the objective rewards only POIs not yet seen.

The desired terminal quaternion $\qvec_\text{des}$ is then the shortest rotation aligning the sensor bore-sight with the line of sight to $l_{k^\star}$, where $l_k$ is the position of the POI, which becomes the terminal pointing of the attitude Quadratic Program (QP)~\eqref{eq:attitudeqp}.

\textit{Continuous Planning.} A trajectory layer converts these targets $c^\star$ and $k^\star$ into references through two convex programs. 

Thrust Planning: Given $c^\star$, the trajectory layer generates a minimum-fuel impulse sequence that transfers the chaser from current PRO to the selected PRO.
\begin{equation}
    \Delta V_\text{seq}
    = \arg\min_{\{\Delta v_k\}}\; \sum_{k=0}^{N-1}\|\Delta v_k\|_1,
    \label{eq:thrustlp}
\end{equation}
subject to thrust limits $\|\Delta v_k\|_\infty \le \Delta v_\text{max}$, and a linearized spherical keep-out constraint resolved by sequential convex programming. 
The solution of~\eqref{eq:thrustlp} is the translational reference $(\pvec_{\text{ref}}, \vvec_{\text{ref}}, \Delta V_{\text{seq}})$, i.e.\ the planned position, velocity, and impulse profile of the transfer. This reference is handed to the thrust controller. 

Attitude Planning : Given $k^\star$, an attitude QP tracks a Spherical Linear Interpolation (SLERP) from the current quaternion $\hat{q}$ to the terminal
pointing $\qvec_\text{des}$ while minimizing total control torque,
\begin{equation}
  \min_{\{q_k,\,\omega_k,\,\tau_k\}}\;
    \Delta t_a \sum_{k=0}^{T_a-1}\|\tau_k\|_2^2,
  \label{eq:attitudeqp}
\end{equation}
subject to linearized attitude dynamics, terminal conditions
$q_{T_a} = \qvec_\text{des}$ and $\omega_{T_a} = 0$, and
per-axis torque and torque-rate limits.
 
\textit{Attitude controller.} The attitude reference $(q_\text{ref}, \omega_\text{ref})$ produced by
\eqref{eq:attitudeqp} is tracked by a robust nonlinear law~\cite{bandyopadhyay2016} that achieves global exponential contraction of the error dynamics under inertia uncertainty,
\begin{equation}
\boldsymbol{\tau} = \hat{\Jmat}\,\dot{\wvec}_r
- \skewm{\hat{\Jmat}\,\hat{\wvec}}\,\wvec_r
- \mathbf{K}_r\,(\hat{\wvec} - \wvec_r),
\label{eq:arch_att_ctrl}
\end{equation}
with reference angular velocity
\begin{equation}
\wvec_r = \mathbf{Z}^{-1}(\hat{\qvec})\,\dot{\qvec}_d
+ \mathbf{Z}^{-1}(\hat{\qvec})\,\boldsymbol{\Lambda}_r\,(\qvec_d - \hat{\qvec}).
\label{eq:arch_omega_ref}
\end{equation}
where $\text{J}$ is the inertial matrix of observer spacecraft, $\hat{\omega}$ is the estimated body angular velocity (chaser), $\hat{\qvec}$ is the estimated attitude of chaser and ${\qvec_d}$ is the desired (reference) attitude quaternion, $\boldsymbol{\Lambda}_r$ is the positive definite convergence rate matrix and sets the rate for the attitude error $(\qvec_d-\hat{\qvec})$ to be zero and $\mathbf{Z}(\hat{\qvec})$ is the kinematic map relating body rates to quaternion rates.  
 
\textit{Thrust controller.} 
The thrust controller is an equivalent optimization problem to the thrust planner except, whereas the thrust planner solves a boundary value problem to the desired PRO $c^*$, the thrust controller tracks the reference $\Delta V_{\text{seq}}$. The thrust controller runs in a model-predictive control fashion at a higher rate. 
\section{Methods}
\label{sec:method}
The complete GNC stack is implemented in NVIDIA IsaacSim with the Lunar Gateway as a USD asset; the chaser carries a single forward-facing camera rendering $256\times256$ RGB frames through the photorealistic ray-traced pipeline. 
All three inspection probabilities share the planner machinery, PRO pool, SCP transfer, attitude QP, controllers, chaser, target, and environment, differing only in the detection probability $\nu_k(x)$ of the inspection transition in Eq.~(1).

\subsection{Geometric baseline $\nu_1$~\cite{nakka2022}}
An inverse-square detection probability gated by a binary visibility test,
\begin{equation}
    \nu_{1,k}(x)=\begin{cases}
        1/d_k(x)^2, & x\in\mathcal{V}_k,\\
        0, & \text{otherwise,}
    \end{cases}
\end{equation}
where $d_k(x)$ is the chaser POI distance and $\mathcal{V}_k$ is the set of physical states at which the $k$th POI is in range, in field of view, at favorable incidence, and unoccluded.

\subsection{Illumination-aware analytic baseline $\nu_2$~\cite{vanwijk2024}}
Following~\cite{vanwijk2024}, a POI is detected only if it is visible, Sun-facing, and adequately lit:
\begin{equation}
\nu_{2,k}(x)=
\begin{cases}
1, & x\in\mathcal{V}_k,\ \hat{N}_k\cdot\hat{L}>0,\ I_k\in[I_{\min},I_{\max}],\\
0, & \text{otherwise,}
\end{cases}
\end{equation}
where $\mathcal{V}_k$ is the set of chaser states $x$ for which POI $k$ lies within the sensor field of view (FOV) and is unoccluded, $\hat{N}_k$ is the POI's surface normal, $\hat{L}$ is the Sun direction, and $I_k$ is its normalized Blinn-Phong illumination. Patches with $I_k$ outside $[I_{\min},I_{\max}]=[0.20,0.83]$ are too dark or too bright to be useful. 
Both baselines are image-blind: they never use the rendered patch.
\subsection{ Learned model $\nu_3$ }
\label{vae_mod}
The per-observation score is the inverse encoder latent variance,
\begin{equation}
\nu_3(s,a)=\frac{1}{1+\sigma_\theta^2(o_k)},
\end{equation}
which consumes the actual rendered image: dark, occluded, or specularly saturated patches yield high variance and a low score; well-resolved, in-distribution patches yield a high score.

\begin{figure}
    \centering
    \includegraphics[width=\columnwidth]{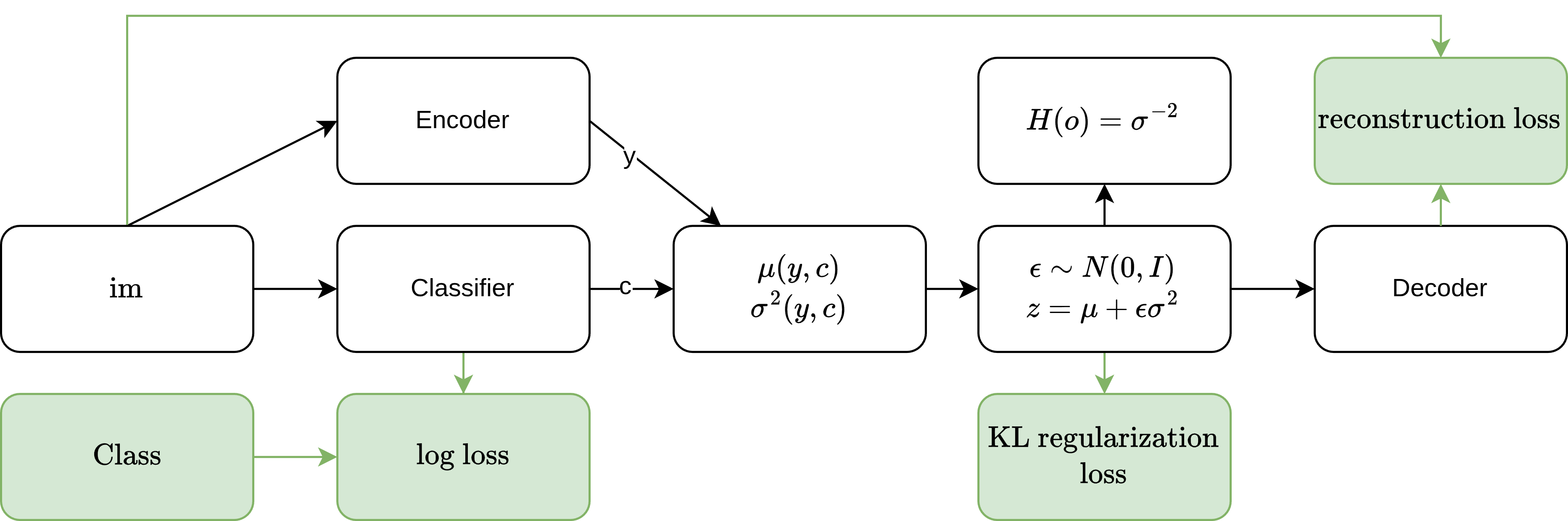}
    \caption{Variational Auto Encoder Architecture adapted from \cite{kingma}. Blocks in green represent the loss functions used in the model. }
    \label{fig:vae}
\end{figure}

\subsubsection{Architecture.} 
The learned inspection probability is a classifier variational autoencoder~\cite{kingma2014vae,kingma2014semi} on local image patches $x\in\mathbb{R}^{3\times64\times64}$. Let $c$ denote the component class of a patch and $z\in\mathbb{R}^{d_z}$, $d_z=32$, a latent code with prior $p(z)=\mathcal{N}(0,I)$. The generative model is $p_\theta(x\mid z,c)\,p(z)$, where the decoder $p_\theta(x\mid z,c)$ reconstructs the patch. 
The inference model consists of a classifier $q_\psi(c\mid x)$, a three-layer MLP, and an encoder
\begin{equation}
    q_\phi(z\mid x,c)=\mathcal{N}\!\big(\mu_\phi(x,c),\,\mathrm{diag}\,\sigma^2_\phi(x,c)\big),
\end{equation} 
The parameters $(\theta,\phi,\psi)$ are trained jointly as described below; the expectation over $z$ is estimated by the reparameterization $z=\mu_\phi+\sigma_\phi\odot\epsilon$, $\epsilon\sim\mathcal{N}(0,I)$. 
At inference time the posterior variance $\sigma^2_\phi(x,c)$ measures how confidently a patch is encoded, and the detection probability $\nu_3$ is derived from it, so that accumulating confident encoded observations increases the reward in Eq.~\eqref{reward}.

\subsubsection{Training.}
The networks are trained by minimizing a weighted negative evidence lower bound~\cite{kingma2014vae,higgins2017beta} augmented with a classification term,
\begin{equation}\label{eq:VAE_loss}
\begin{split}
\mathcal{L} = {}& \underbrace{\lVert x - \hat{x} \rVert_{2}^{2}}_{\mathcal{L}_{\text{rec}}}
+ \beta \, \underbrace{D_{\text{KL}}\!\bigl(q_\phi(z \mid x, c) \,\Vert\, p(z)\bigr)}_{\mathcal{L}_{\text{KL}}} \\
&+ \lambda_{\text{cls}} \, \underbrace{\mathrm{CE}\!\bigl(q_\psi(c \mid x),\; c\bigr)}_{\mathcal{L}_{\text{cls}}},
\end{split}
\end{equation} where $\hat{x}$ is the decoder reconstruction from a single re-parameterized sample $z\sim q_\phi(z\mid x,c)$, so that $\mathcal{L}_\text{rec}$ is the negative Gaussian log-likelihood up to a constant; $\mathcal{L}_\text{KL}$ regularizes the class-conditional posterior toward the prior; and $\mathcal{L}_\text{cls}$ is a label-smoothed cross-entropy between the classifier output and the ground-truth class $c$. 
 The dataset is constructed with the class taxonomy is derived from the Lunar Gateway Universal Scene Description (USD) hierarchy (e.g., solar panel, shutter, airlock, habitat module, thermal radiator etc.); patches are labeled by a $3\times3$ raycast majority vote and split by PRO index 80/20, a stricter generalization test than a patch-level split. 

\section{Experiments}
\label{experiments}

A quantitative and qualitative analysis are presented to test the three different inspection probabilities. 

\subsection{Experimental Setup}
The experimental set up in IsaacSim used open source NASA models~\cite{nasa3dresources} of the Lunar Gateway for the target and a general satellite model adapted from CloudSat models for the chaser along with an Earth, Moon and Sun module. 

\subsection{Quantitative}

\begin{figure}[h!]
\centering
\includegraphics[width=1\columnwidth]{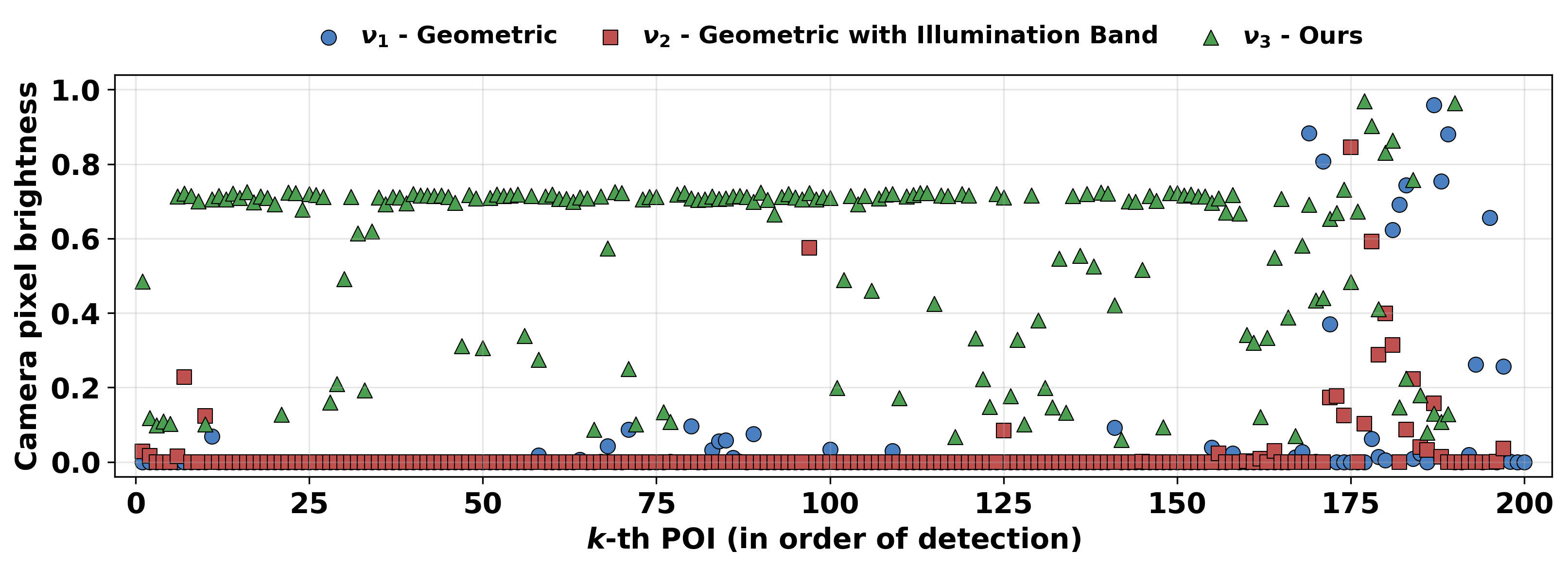}
\includegraphics[width=\columnwidth]{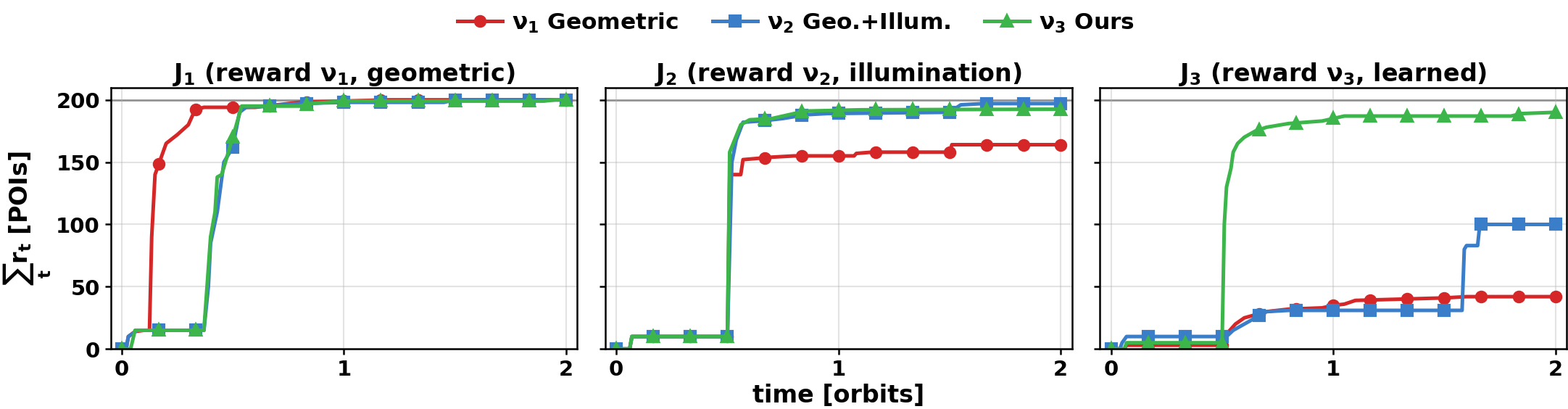}

\caption{Observations of POIs under the three inspection probabilities. 
In graph 1, the $x$-axis is the POIs at first observation; the $y$-axis is direct camera pixel brightness from the photorealistic simulator. Graph 2 is a combination of the three functions maximizing reward under $\nu_1$, $\nu_2$ and $\nu_3$ functions for target's orbits around the earth.
}
\label{fig:200}
\end{figure}

A quantitative analysis is presented to test the three detection probability functions $\nu_1$, $\nu_2$ and $\nu_3$ under identical initial conditions. The three methods are evaluated on two metrics: (i) camera pixel intensity rendered by the photorealistic simulator when each POI is first marked observed, and (ii) reward maximization across all three inspection probabilities.
The analytical methods $\nu_1$ and $\nu_2$ are largely uninformative with respect to image quality. Under $\nu_1$, $83.5\%$ of POI's are admitted at zero brightness (mean $0.04$, median $0.00$), and only $6.0\%$ of observations exceed $0.25$. Under $\nu_2$, the illumination band does not improve this behavior: $87.3\%$ of POI's are still admitted at zero brightness, for an overall mean brightness of $0.02$ (median $0.00$, $3.0\%$ above $0.25$). In contrast, $\nu_3$ admits every POI at nonzero brightness, with $62.6\%$ of observations concentrated between $0.65$ and $0.75$, for a mean brightness of $0.57$ (median $0.71$, $81.6\%$ above $0.25$, $96.3\%$ above $0.1$). The low-brightness observations $\nu_3$ does make occur only when the POI class is correctly identified. These results demonstrate that the proposed method $\nu_3$ overcomes the limitations of analytical models by operating directly on the rendered image.
To assess each policy's behavior (since each optimizes a different objective), we evaluate every policy $\pi_i$ under every reward $\nu_j$, reporting $J_j(\pi_i)$ over $2$ orbits in Fig.~\ref{fig:200}. Under the purely geometric reward $J_1$, all three policies reach the full $200$ POI's. $\pi_1$ converges fastest, reaching $\approx 193$ POI's within $0.33$ orbits versus $\approx 0.55$ orbits for $\pi_2$ and $\pi_3$, indicating that the geometric criterion is overconfident in its inspection and admits false positives. Under $J_2$, $\pi_1$ attains $164$ POI's, while $\pi_3$ attains $193$ and $98\%$ of the $197$ achieved by $\pi_2$ on its own reward. The reverse does not hold: under $J_3$, $\pi_2$ attains only $100$ POI's ($53\%$ of $\pi_3$) and $\pi_1$ only $42$ ($22\%$), whereas $\pi_3$ reaches $190$, with $\approx 185$ of these inspected within the first orbit. The learned reward therefore largely subsumes the analytical illumination criterion, while the analytical criterion does not capture what $\nu_3$ requires. As Fig.~\ref{fig:200} shows, many POI's that $\nu_2$ credits as illuminated are in fact rendered dark in the camera image, primarily because $\nu_2$ does not account for eclipsing or self-shadowing.

\textit{Case where the geometric baselines are adequate:}
On well-illuminated, geometrically simple regions under favorable Sun geometry, the analytical criteria are adequate proxies for image quality: the $12$ POI's that $\nu_1$ observes above a brightness of $0.25$ are captured at a mean brightness of $0.66$, exceeding the $0.50$ achieved by $\nu_3$ on the same POI's.

\begin{table}[t]
\centering
\caption{Inspection reward $J_j(\pi_i)$ across orbital regimes: POIs detected under reward $\nu_j$ over 2 orbits of the target.}
\label{tab:orbits}
\footnotesize
\setlength{\tabcolsep}{4pt}
\begin{tabular}{llcccc}
\toprule
Orbit & Method & $J_1$ & $J_2$ & $J_3$ & Gap \\
\midrule
\multirow{3}{*}{\shortstack[l]{LEO\\\scriptsize 400\,km}}
 & $\nu_1$ Geometric    & 200 & 76  & 82           &       \\
 & $\nu_2$ Geo.+Illum.  & 200 & 200 & 22           &       \\
 & $\nu_3$ Ours         & 200 & 196 & \textbf{154} & $+72$ \\
\midrule
\multirow{3}{*}{\shortstack[l]{MEO\\\scriptsize 20\,200\,km}}
 & $\nu_1$ Geometric    & 200 & 87  & 92           &       \\
 & $\nu_2$ Geo.+Illum.  & 200 & 200 & 23           &       \\
 & $\nu_3$ Ours         & 200 & 199 & \textbf{150} & $+58$ \\
\midrule
\multirow{3}{*}{\shortstack[l]{GEO\\\scriptsize 35\,786\,km}}
 & $\nu_1$ Geometric    & 200 & 88  & 97           &       \\
 & $\nu_2$ Geo.+Illum.  & 200 & 200 & 22           &       \\
 & $\nu_3$ Ours         & 199 & 194 & \textbf{160} & $+63$ \\
\bottomrule
\end{tabular}
\end{table}

We analyze performance across multiple orbital regimes and report results in Table~\ref{tab:orbits}. When evaluated under objectives other than its own, $\nu_3$ exhibits negligible degradation: averaged across the MEO and GEO regimes, it retains 99.8\% of $\nu_1$'s performance under $J_1$ (199.5 vs.\ 200) and 98.3\% of $\nu_2$'s performance under $J_2$ (196.5 vs.\ 200). In contrast, the baseline policies degrade substantially when evaluated under the learned reward $\nu_3$: $\nu_1$ attains only 61.0\% of the $J_3$ achieved by $\nu_3$ (94.5 vs.\ 155), and $\nu_2$ attains only 14-15\% (22-23 vs.\ 150-160).

\begin{figure}[t]
\centering
\newcommand{\gridw}{0.30\columnwidth}
\newcommand{\labw}{0.06\columnwidth}
\newcommand{\cellh}{1.6cm}
\newcommand{\rowlab}[1]{\rotatebox[origin=c]{90}{\textbf{#1}}}
\newcommand{\cell}[1]{%
  \begin{subfigure}[c]{\gridw}\centering
    \includegraphics[width=\linewidth,height=\cellh,keepaspectratio]{#1}%
  \end{subfigure}}

\begin{subfigure}[c]{\labw}\centering\phantom{x}\end{subfigure}\hfill
\begin{subfigure}[c]{\gridw}\centering\footnotesize\boldmath\textbf{$t=5$ s}\end{subfigure}\hfill
\begin{subfigure}[c]{\gridw}\centering\footnotesize\boldmath\textbf{$t=100$ s}\end{subfigure}\hfill
\begin{subfigure}[c]{\gridw}\centering\footnotesize\boldmath\textbf{$t=800$ s}\end{subfigure}

\vspace{2pt}
\begin{subfigure}[c]{\labw}\centering\rowlab{$\boldsymbol{\nu_1}$}\end{subfigure}\hfill
\cell{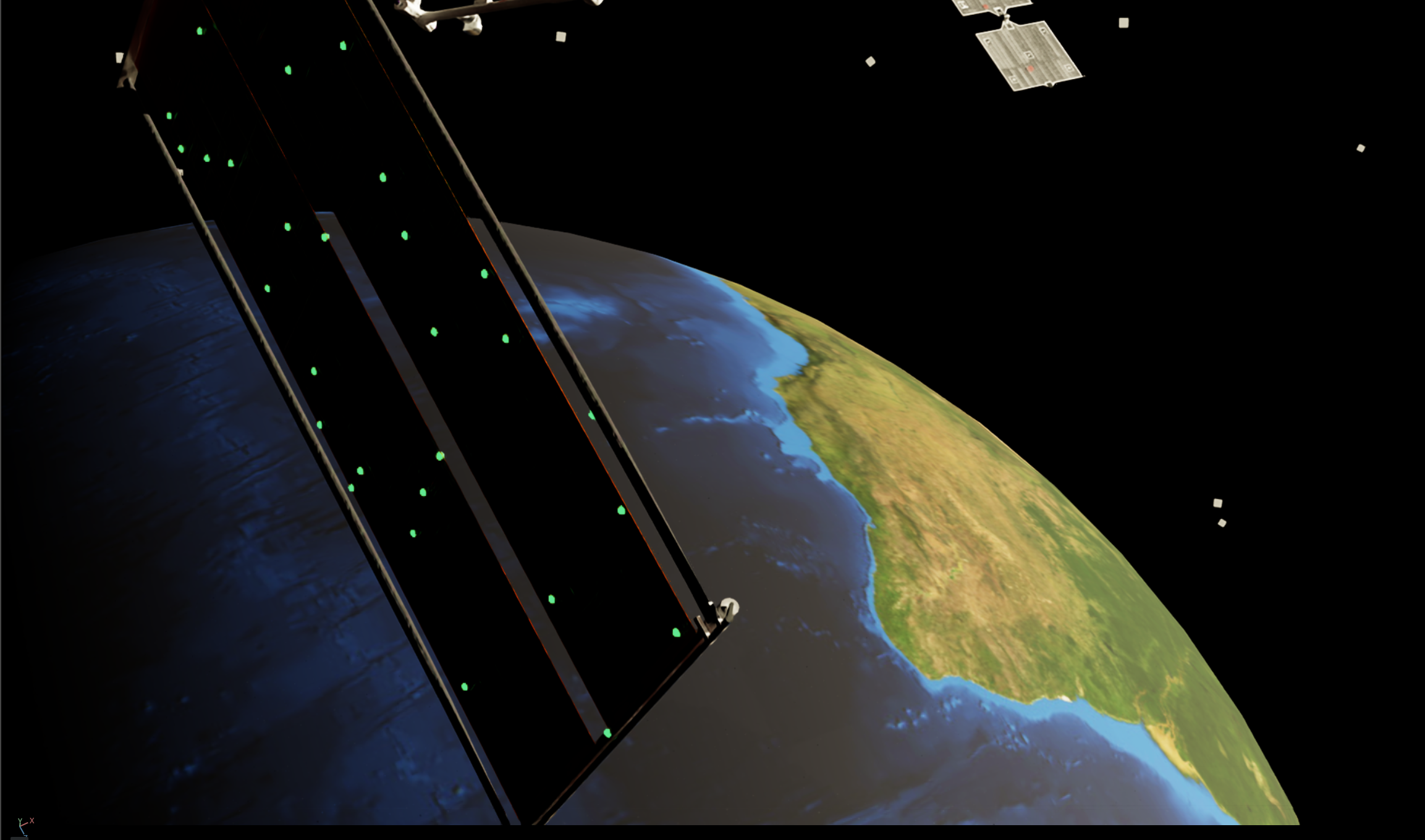}\hfill
\cell{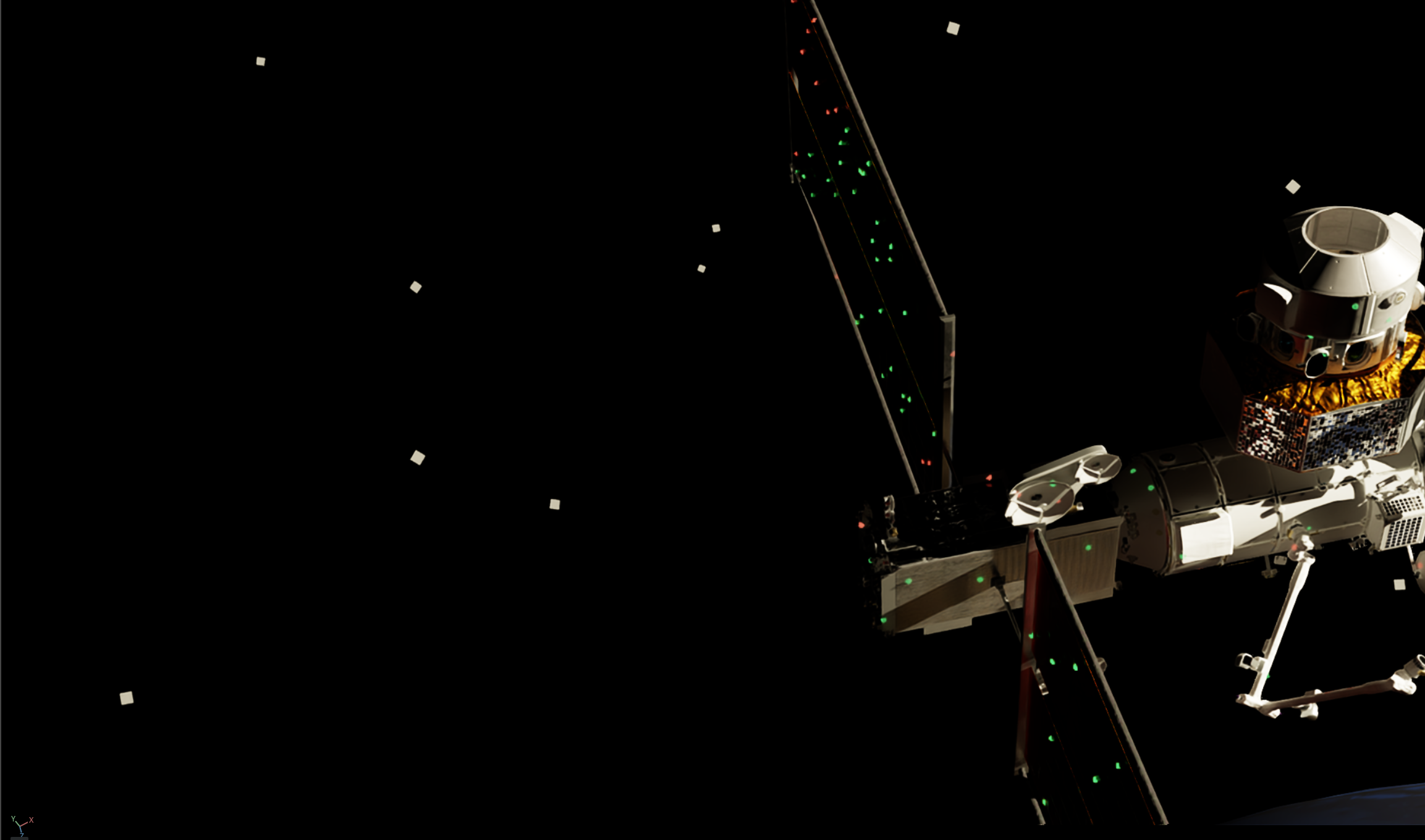}\hfill
\cell{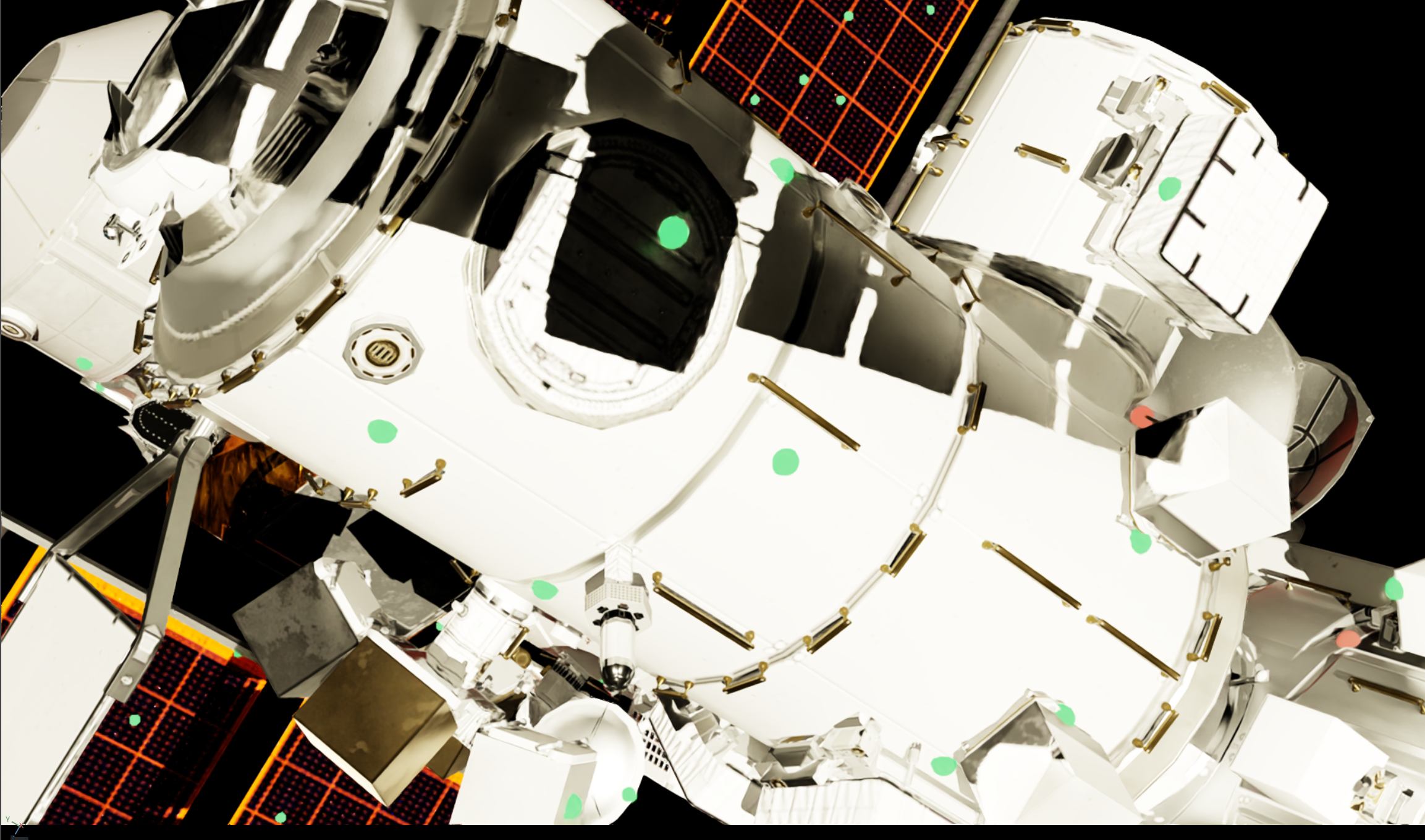}

\vspace{2pt}
\begin{subfigure}[c]{\labw}\centering\rowlab{$\boldsymbol{\nu_2}$}\end{subfigure}\hfill
\cell{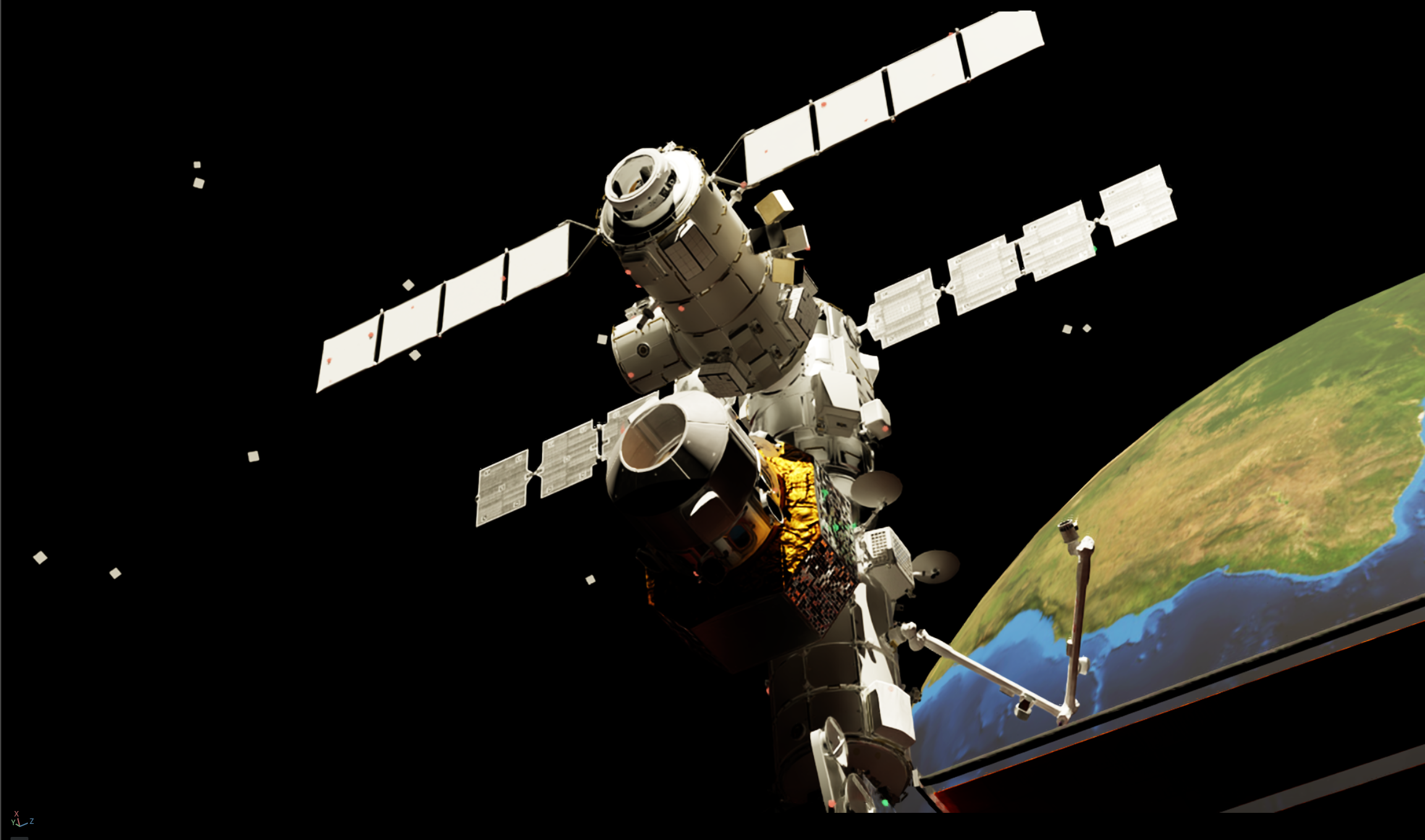}\hfill
\cell{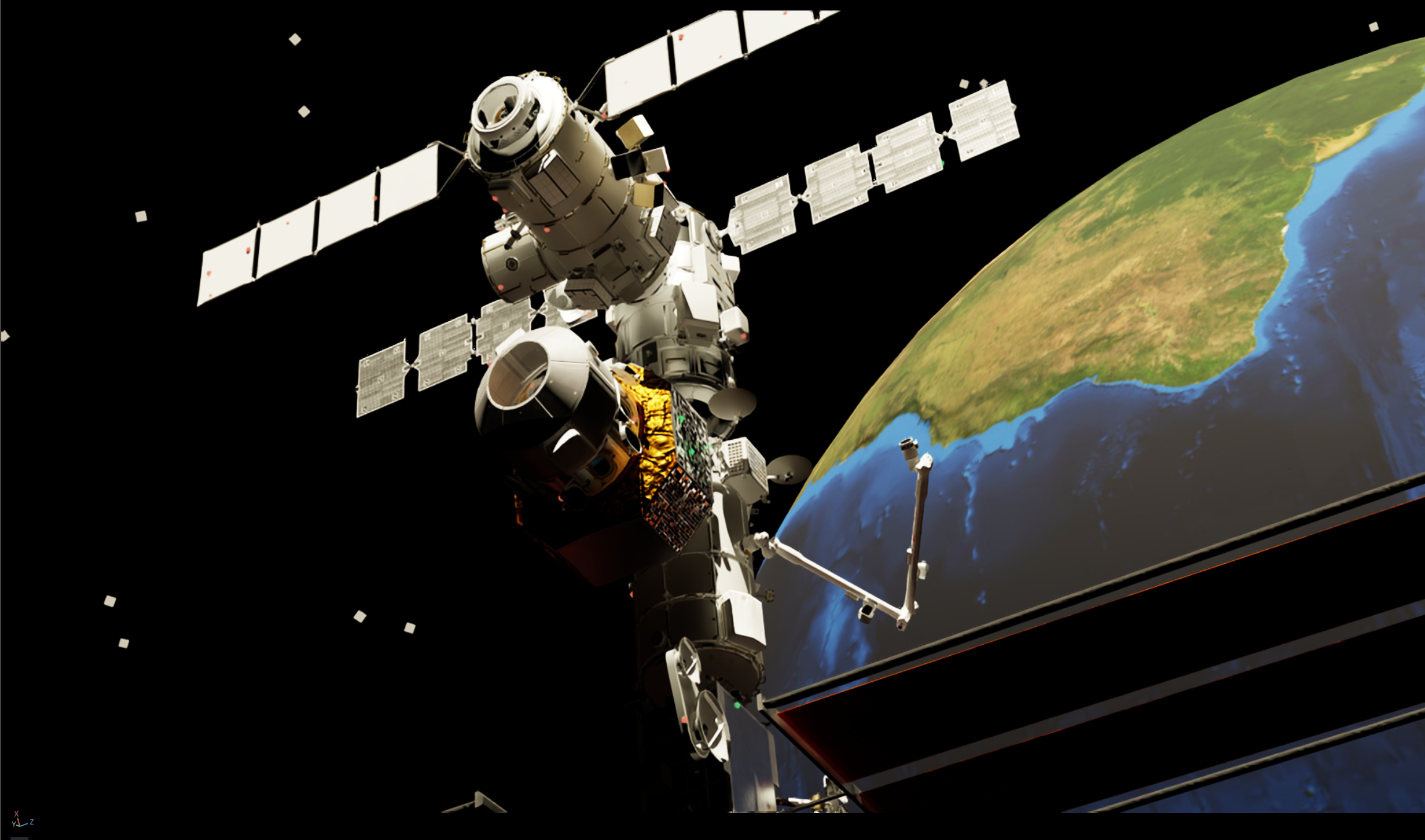}\hfill
\cell{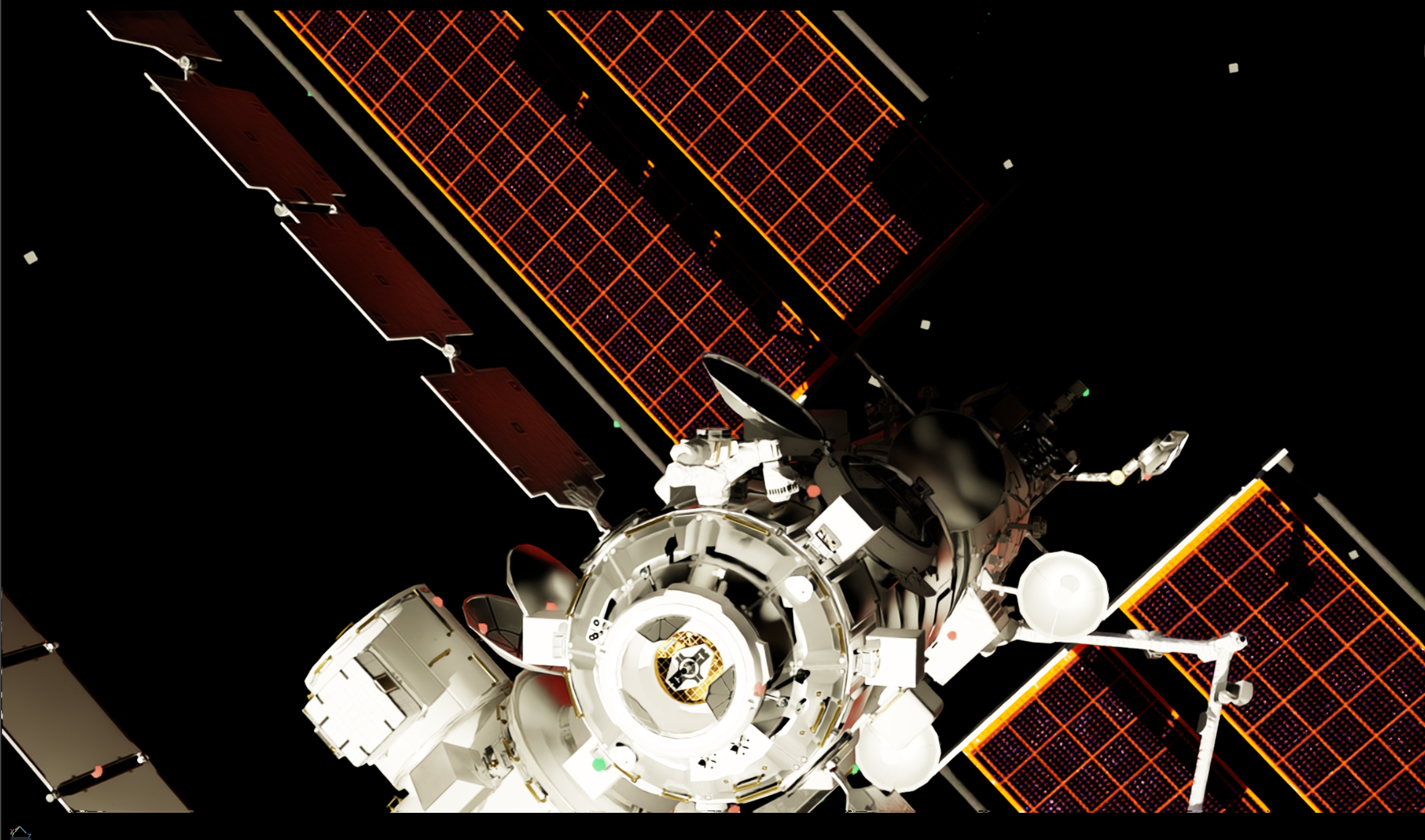}

\vspace{2pt}
\begin{subfigure}[c]{\labw}\centering\rowlab{$\boldsymbol{\nu_3}$}\end{subfigure}\hfill
\cell{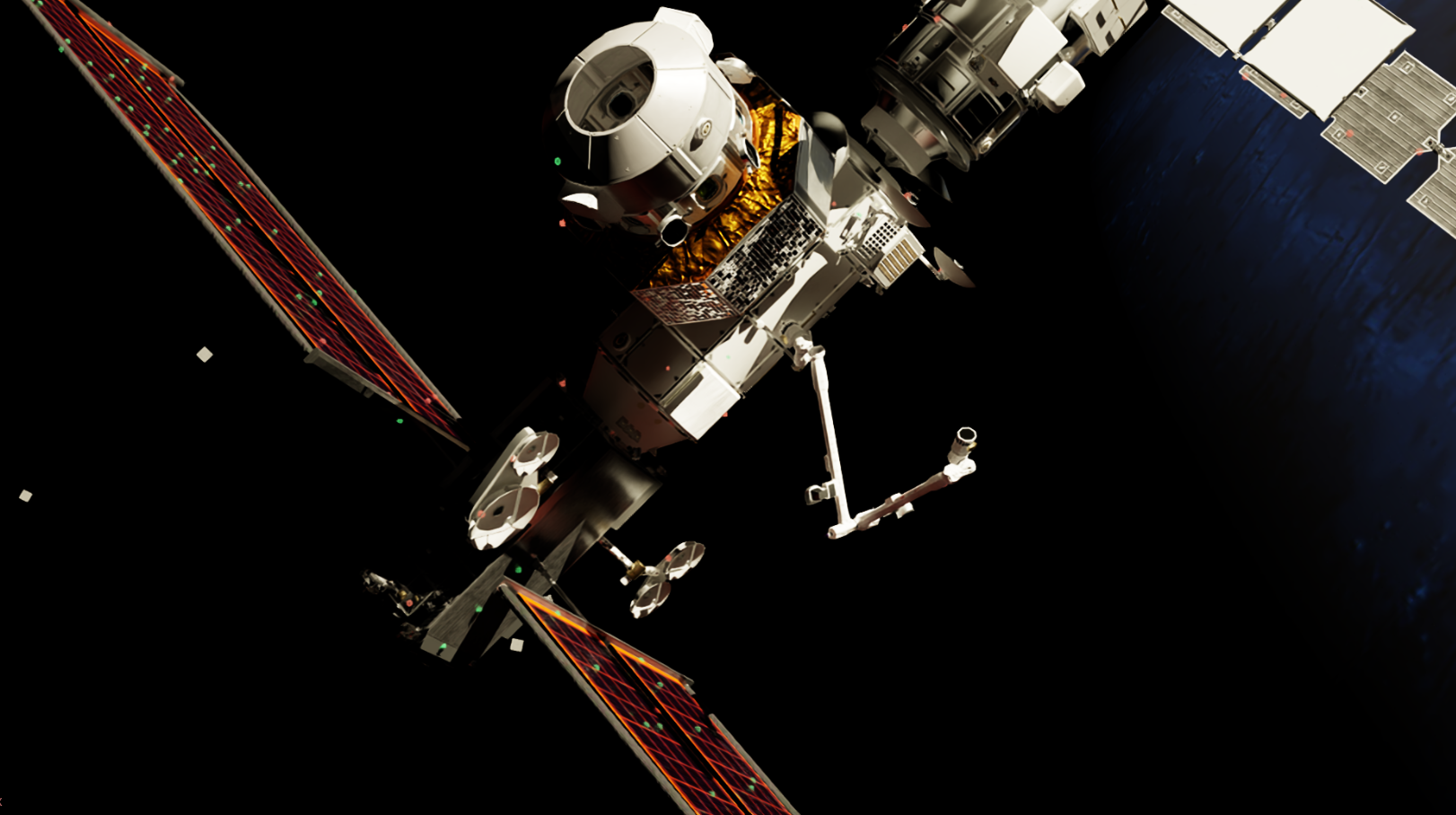}\hfill
\cell{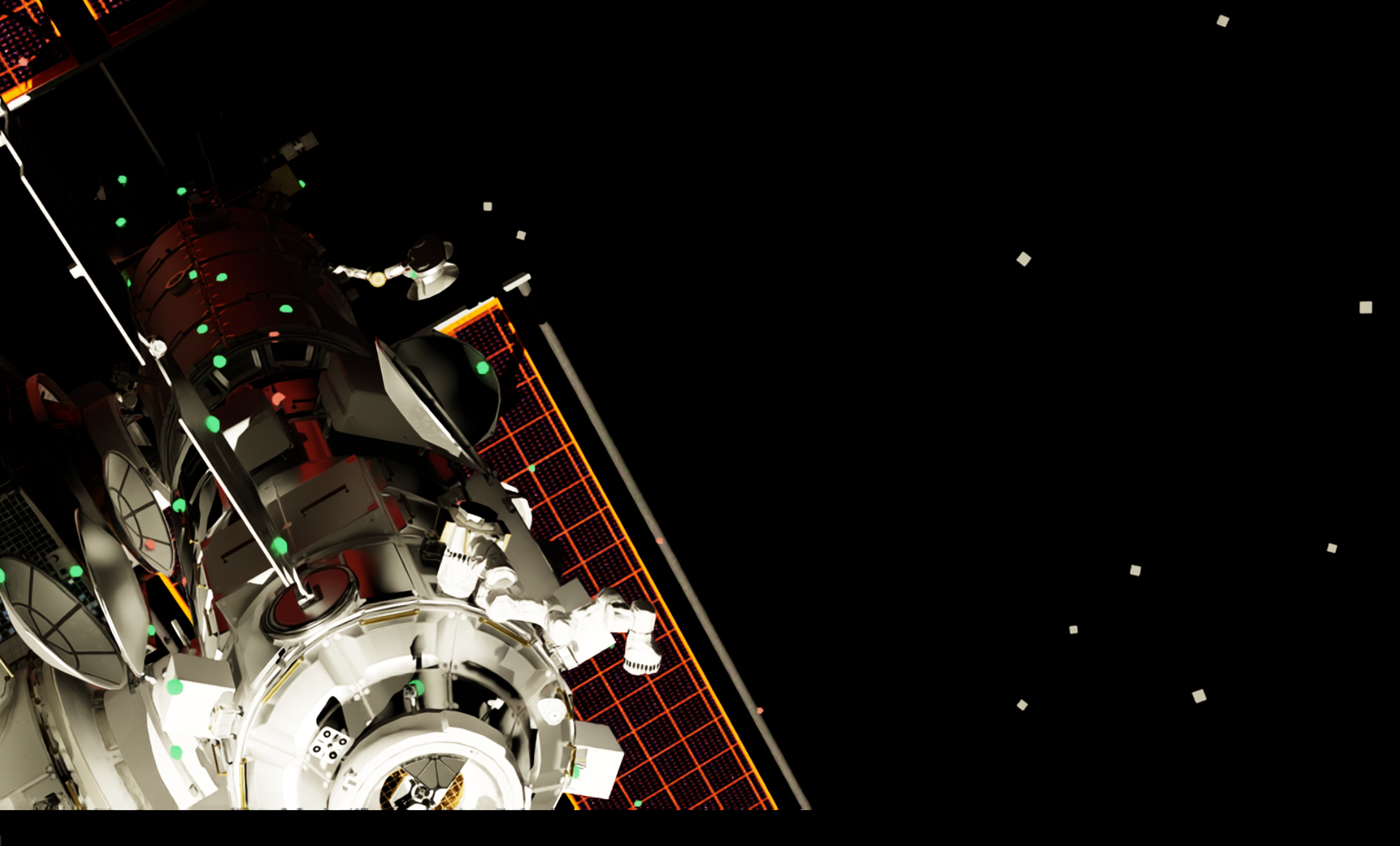}\hfill
\cell{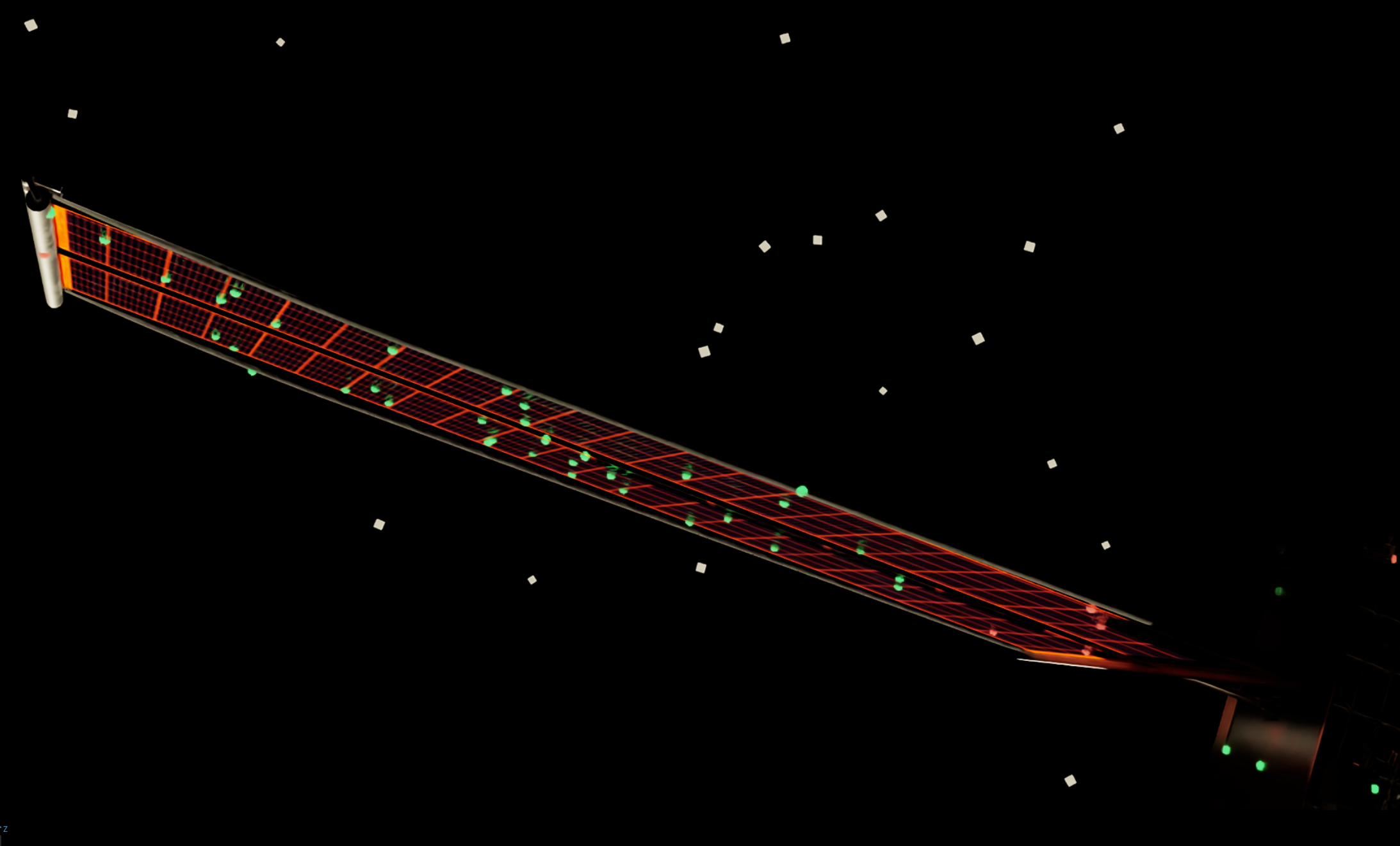}

\caption{Qualitative comparison of the three inspection probabilities during inspection. Rows show $\nu_1$ (geometric), $\nu_2$ (geometric with illumination band), and $\nu_3$ (ours); columns show the camera view at $t=5$, $100$, and $800$~s.}
\label{fig:poi_sets}
\end{figure}

\subsection{Qualitative analysis}
A qualitative analysis is performed by rendering equidistant time steps of the inspection mission defined by the various inspection probabilities. These images are captured by the camera sensor onboard the chaser spacecraft. In Fig ~\ref{fig:poi_sets} it is evidently seen that $\nu_1$ conclusively observes POI's placed on the target at t=5 secs even though the module is in complete darkness. This is seen through the red dots turning green and are considered false positives. It is nearly impossible to distinguish the features on the module and yet the model declares them as observed. Contrarily, $\nu_2$ picks viewing angles to observe POI's under better lighting conditions. However as seen at t=5 secs and t=100 secs (making two similar orbital passes), it rejects POI's due to its conservative illumination band limits resulting in false negatives and also fails to account for self shadows.
The learned model $\nu_3$ overcomes the limitations of the geometric models by observing the POI's under good lighting conditions and proceeding to also admit dim POI's placed between modules. 

\section{Conclusion and Future Work}
We present an illumination-aware active-perception framework that replaces geometric visibility with a learned, image-based detection score from a classifier VAE in a complete GNC system. By evaluating information from the captured image directly, the framework remains accurate under harsh lighting conditions for complex structures in orbit and hence is a robust alternative to analytical models for orbital inspection. A limitation of using classifier model uncertainty for inspection is, the difficulty in determining whether high uncertainty reflects genuine ambiguity in the image itself or is instead a consequence of the model's architecture and/or training limitations. Future work would extend to fault detection, where the learned model can differentiate between healthy components and broken modules while maintaining confidence of these observations over uncertainty. We will also test our solution on a physical robotic free-flyer hardware under harsh lighting conditions to further validate the model.

\bibliography{references.bib}

@inproceedings{isler2016information,
  author    = {Isler, Stefan and Sabzevari, Reza and Delmerico, Jeffrey and Scaramuzza, Davide},
  title     = {An Information Gain Formulation for Active Volumetric 3D Reconstruction},
  booktitle = {IEEE International Conference on Robotics and Automation (ICRA)},
  pages     = {3477--3484}, year = {2016},
  doi       = {10.1109/ICRA.2016.7487527}
}

@article{delmerico2018comparison,
  author  = {Delmerico, Jeffrey and Isler, Stefan and Sabzevari, Reza and Scaramuzza, Davide},
  title   = {A Comparison of Volumetric Information Gain Metrics for Active 3D Object Reconstruction},
  journal = {Autonomous Robots},
  volume  = {42}, number = {2}, pages = {197--208}, year = {2018},
  doi     = {10.1007/s10514-017-9634-0}
}

@article{mendoza2020nbvnet,
  author  = {Mendoza, Miguel and Vasquez-Gomez, J. Irving and Taud, Hind and Sucar, L. Enrique and Reta, Carolina},
  title   = {Supervised Learning of the Next-Best-View for 3D Object Reconstruction},
  journal = {Pattern Recognition Letters},
  volume  = {133}, pages = {224--231}, year = {2020},
  doi     = {10.1016/j.patrec.2020.02.024}
}

@inproceedings{zeng2020pcnbv,
  author    = {Zeng, Rui and Zhao, Wang and Liu, Yong-Jin},
  title     = {{PC-NBV}: A Point Cloud Based Deep Network for Efficient Next Best View Planning},
  booktitle = {IEEE/RSJ International Conference on Intelligent Robots and Systems (IROS)},
  pages     = {7050--7057}, year = {2020},
  doi       = {10.1109/IROS45743.2020.9340916}
}

@inproceedings{peralta2020nbvpolicy,
  author    = {Peralta, Daryl and Casimiro, Joel and Nilles, Aldrin Michael and Aguilar, Justine Aletta and Atienza, Rowel and Cajote, Rhandley},
  title     = {Next-Best View Policy for 3D Reconstruction},
  booktitle = {Computer Vision -- ECCV 2020 Workshops},
  series    = {Lecture Notes in Computer Science},
  volume    = {12538},
  publisher = {Springer},
  year      = {2020},
  doi       = {10.1007/978-3-030-66823-5_33}
}

@inproceedings{pan2022activenerf,
  author    = {Pan, Xuran and Lai, Zihang and Song, Shiji and Huang, Gao},
  title     = {{ActiveNeRF}: Learning Where to See with Uncertainty Estimation},
  booktitle = {Computer Vision -- ECCV 2022},
  series    = {Lecture Notes in Computer Science},
  volume    = {13693},
  pages     = {230--246},
  publisher = {Springer},
  year      = {2022},
  doi       = {10.1007/978-3-031-19827-4_14}
}

@article{lee2022uncertainty,
  author  = {Lee, Soomin and Chen, Le and Wang, Jiahao and Liniger, Alexander and Kumar, Suryansh and Yu, Fisher},
  title   = {Uncertainty Guided Policy for Active Robotic 3D Reconstruction Using Neural Radiance Fields},
  journal = {IEEE Robotics and Automation Letters},
  volume  = {7}, number = {4}, pages = {12070--12077}, year = {2022},
  doi     = {10.1109/LRA.2022.3213827}
}

@inproceedings{jin2023neunbv,
  author    = {Jin, Liren and Chen, Xieyuanli and R{\"u}ckin, Julius and Popovi{\'c}, Marija},
  title     = {{NeU-NBV}: Next Best View Planning Using Uncertainty Estimation in Image-Based Neural Rendering},
  booktitle = {IEEE/RSJ International Conference on Intelligent Robots and Systems (IROS)},
  pages     = {11305--11312}, year = {2023},
  doi       = {10.1109/IROS55552.2023.10342512}
}

@article{ran2023neurar,
  author  = {Ran, Yunlong and Zeng, Jing and He, Shibo and Chen, Jiming and Li, Lincheng and Chen, Yingfeng and Lee, Gim Hee and Ye, Qi},
  title   = {{NeurAR}: Neural Uncertainty for Autonomous 3D Reconstruction with Implicit Neural Representations},
  journal = {IEEE Robotics and Automation Letters},
  volume  = {8}, number = {2}, pages = {1125--1132}, year = {2023},
  doi     = {10.1109/LRA.2023.3235686}
}

@inproceedings{jiang2023fisherrf,
  author    = {Jiang, Wen and Lei, Boshu and Daniilidis, Kostas},
  title     = {{FisherRF}: Active View Selection and Uncertainty Quantification for Radiance Fields Using Fisher Information},
  booktitle = {Computer Vision -- ECCV 2024},
  series    = {Lecture Notes in Computer Science},
  publisher = {Springer},
  year      = {2024},
  note      = {arXiv:2311.17874}
}

@misc{costante2018perception,
  author = {Costante, Gabriele and Forster, Christian and Delmerico, Jeffrey and Valigi, Paolo and Scaramuzza, Davide},
  title  = {Perception-Aware Path Planning},
  howpublished = {arXiv preprint arXiv:1605.04151},
  year   = {2016}
}

@inproceedings{zhang2019fisher,
  author    = {Zhang, Zichao and Scaramuzza, Davide},
  title     = {Beyond Point Clouds: Fisher Information Field for Active Visual Localization},
  booktitle = {IEEE International Conference on Robotics and Automation (ICRA)},
  pages     = {5986--5992}, year = {2019},
  doi       = {10.1109/ICRA.2019.8793405}
}

@article{sabatini2020,
  author  = {Sabatini, M. and Volpe, R. and Palmerini, G. B.},
  title   = {Centralized Visual Based Navigation and Control of a Swarm of Satellites for On-Orbit Servicing},
  journal = {Acta Astronautica},
  volume  = {171},
  pages   = {323-334},
  month   = jun,
  year    = {2020}
}

@misc{NVIDIA_Isaac_Sim,
  author       = {{NVIDIA}},
  title        = {Isaac {Sim}},
  howpublished = {\url{https://developer.nvidia.com/isaac/sim}},
  year         = {2024},
  note         = {Accessed: Sep. 24, 2026}
}

@article{vanwijk2024,
  author  = {van Wijk, D. E. J. and Dunlap, K. and Majji, M. and Hobbs, K.},
  title   = {Safe Spacecraft Inspection via Deep Reinforcement Learning and Discrete Control Barrier Functions},
  journal = {Journal of Aerospace Information Systems},
  volume  = {21},
  number  = {12},
  pages   = {996-1013},
  month   = dec,
  year    = {2024},
  doi     = {10.2514/1.I011391}
}

@misc{aurand2023,
  author        = {Aurand, J. and Cutlip, S. and Lei, H. and Lang, K. and Phillips, S.},
  title         = {Exposure-Based Multi-Agent Inspection of a Tumbling Target Using Deep Reinforcement Learning},
  year          = {2023},
  eprint        = {2302.14188},
  archivePrefix = {arXiv},
  primaryClass  = {cs.LG}
}

@inproceedings{lauinger,
  author    = {Lauinger, T. and Ulrich, S.},
  title     = {Path Planning for Optimal Coverage of Orbiting Space Structures Using Lissajous Curves},
  booktitle = {AAS/AIAA Space Flight Mechanics Meeting},
  month     = jan,
  year      = {2025}
}

@article{nakka2022,
  author  = {Nakka, Y. K. and Honig, W. and Choi, C. and Harvard, A. and Rahmani, A. and Chung, S.-J.},
  title   = {Information-Based Guidance and Control Architecture for Multi-Spacecraft On-Orbit Inspection},
  journal = {Journal of Guidance, Control, and Dynamics},
  volume  = {45},
  pages   = {1184-1201},
  year    = {2022}
}

@inproceedings{stephenson2025,
  author    = {Stephenson, M. and Schaub, H.},
  title     = {Safe, Autonomous Multi-Agent Inspection of Space Objects Leveraging Relative Orbit Dynamics},
  booktitle = {Advanced Maui Optical and Space Surveillance Technologies Conference},
  year      = {2025}
}

@article{bandyopadhyay2016,
  author    = {Bandyopadhyay, Saptarshi and Chung, Soon-Jo and Hadaegh, Fred Y.},
  title     = {Nonlinear Attitude Control of Spacecraft with a Large Captured Object},
  journal   = {Journal of Guidance, Control, and Dynamics},
  volume    = {39},
  number    = {4},
  pages     = {754-769},
  month     = apr,
  year      = {2016},
  doi       = {10.2514/1.G001341}
}

@article{malyuta,
  author  = {Malyuta, D. and Yu, Y. and Elango, P. and A{\c{c}}{\i}kme{\c{s}}e, B.},
  title   = {Advances in Trajectory Optimization for Space Vehicle Control},
  journal = {Annual Reviews in Control},
  volume  = {52},
  pages   = {282-315},
  year    = {2021},
  doi     = {10.1016/j.arcontrol.2021.04.013}
}

@inproceedings{kingma,
  author    = {Kingma, Diederik P. and Welling, Max},
  title     = {Auto-Encoding Variational Bayes},
  booktitle = {2nd International Conference on Learning Representations (ICLR)},
  address   = {Banff, AB, Canada},
  year      = {2014},
  eprint    = {1312.6114},
  archivePrefix = {arXiv},
  primaryClass  = {stat.ML}
}

@inproceedings{dunlap20242024,
  author    = {Dunlap, Kyle and Bennett, Kochise and van Wijk, David and Hamilton, Nathaniel and Hobbs, Kerianne},
  title     = {Run Time Assured Reinforcement Learning for Six Degree-of-Freedom Spacecraft Inspection},
  booktitle = {AIAA Aviation Forum and ASCEND},
  year      = {2024},
  month     = jul,
  doi       = {10.2514/6.2024-4935},
  note      = {AIAA 2024-4935}
}

@inproceedings{ravaioli2023,
  author    = {Ravaioli, U. J. and Dunlap, Kyle and Hobbs, Kerianne},
  title     = {A Universal Framework for Generalized Run Time Assurance with {JAX} Automatic Differentiation},
  booktitle = {2023 American Control Conference (ACC)},
  pages     = {4264-4269},
  year      = {2023},
  doi       = {10.23919/ACC55779.2023.10156439}
}

@article{shi2025,
  author  = {Shi, X. and others},
  title   = {Safety-Critical Control for Multiple Spacecraft Close-Proximity Inspection},
  journal = {Aerospace Science and Technology},
  year    = {2025},
  doi     = {10.1016/j.ast.2025.10028}
}

@misc{wijayatunga2026,
  author        = {Wijayatunga, Minduli C. and Linares, Richard and Armellin, Roberto},
  title         = {Meta-Reinforcement Learning for Robust and Non-greedy Control Barrier Functions in Spacecraft Proximity Operations},
  year          = {2026},
  eprint        = {2602.07335},
  archivePrefix = {arXiv},
  primaryClass  = {eess.SY}
}

@misc{dunlap2024uav,
  author        = {Dunlap, Kyle and Hamilton, Nathaniel and Lippay, Zachary and Shubert, Matthew and Phillips, Sean and Hobbs, Kerianne L.},
  title         = {Demonstrating Reinforcement Learning and Run Time Assurance for Spacecraft Inspection Using Unmanned Aerial Vehicles},
  year          = {2024},
  eprint        = {2405.06770},
  archivePrefix = {arXiv},
  primaryClass  = {eess.SY}
}

@misc{starfish2025,
  author       = {{Starfish Space}},
  title        = {{Starfish Space Completes Autonomous Rendezvous and Proximity Mission in LEO with Impulse Space}},
  year         = {2025},
  month        = dec,
  note         = {Accessed: 2026-06-19}
}

@misc{astroscale2026,
  author       = {{Astroscale}},
  title        = {{Astroscale's ADRAS-J Mission Completes Operations, Begins Deorbit}},
  year         = {2026},
  month        = mar,
  note         = {Accessed: 2026-06-19}
}

@article{fourie2014,
  author  = {Fourie, Dehann and Tweddle, Brent E. and Ulrich, Steve and Saenz-Otero, Alvar},
  title   = {Flight Results of Vision-Based Navigation and Control for Autonomous Spacecraft Inspection of an Unknown Object},
  journal = {Journal of Spacecraft and Rockets},
  volume  = {51},
  number  = {6},
  pages   = {2016-2026},
  year    = {2014},
  doi     = {10.2514/1.A32813}
}

@book{puterman,
author = {Puterman, Martin L.},
title = {Markov Decision Processes: Discrete Stochastic Dynamic Programming},
year = {1994},
isbn = {0471619779},
publisher = {John Wiley \& Sons, Inc.},
address = {USA},
edition = {1st}
}

@article{KL,
author = {I. Csiszar},
title = {{$I$-Divergence Geometry of Probability Distributions and Minimization Problems}},
volume = {3},
journal = {The Annals of Probability},
number = {1},
publisher = {Institute of Mathematical Statistics},
pages = {146 -- 158},
year = {1975},
doi = {10.1214/aop/1176996454},
URL = {https://doi.org/10.1214/aop/1176996454}
}

@misc{nasa3dresources,
  author       = {{NASA}},
  title        = {{3D} {Resources}},
  howpublished = {\url{https://science.nasa.gov/3d-resources/}},
  year         = {2026},
  note         = {NASA Science. Accessed: 2026-06-24}
}

@article{AILOR2022449,
title = {Protecting the LEO environment},
journal = {Journal of Space Safety Engineering},
volume = {9},
number = {3},
pages = {449-454},
year = {2022},
issn = {2468-8967},
doi = {https://doi.org/10.1016/j.jsse.2022.07.001},
url = {https://www.sciencedirect.com/science/article/pii/S2468896722000702},
author = {William Ailor}
}

@article{clohessy1960,
  author  = {Clohessy, W. H. and Wiltshire, R. S.},
  title   = {Terminal Guidance System for Satellite Rendezvous},
  journal = {Journal of the Aerospace Sciences},
  volume  = {27},
  number  = {9},
  pages   = {653--658},
  year    = {1960},
  doi     = {10.2514/8.8704}
}

@inproceedings{kingma2014vae,
  title     = {Auto-Encoding Variational {B}ayes},
  author    = {Kingma, Diederik P. and Welling, Max},
  booktitle = {International Conference on Learning Representations (ICLR)},
  year      = {2014}
}

@inproceedings{kingma2014semi,
  title     = {Semi-supervised Learning with Deep Generative Models},
  author    = {Kingma, Diederik P. and Rezende, Danilo J. and Mohamed, Shakir and Welling, Max},
  booktitle = {Advances in Neural Information Processing Systems (NeurIPS)},
  volume    = {27},
  year      = {2014}
}

@inproceedings{higgins2017beta,
  title     = {beta-{VAE}: Learning Basic Visual Concepts with a Constrained Variational Framework},
  author    = {Higgins, Irina and Matthey, Loic and Pal, Arka and Burgess, Christopher and Glorot, Xavier and Botvinick, Matthew and Mohamed, Shakir and Lerchner, Alexander},
  booktitle = {International Conference on Learning Representations (ICLR)},
  year      = {2017}
}

\end{document}